%% file: main.tex
\documentclass{article}  % For LaTeX2e
\usepackage[dvipsnames]{xcolor}

\usepackage{iclr2025_conference,times}

\input{math_commands.tex}

\usepackage{hyperref}
\usepackage{caption}  
\usepackage{mathtools}
\usepackage{url}
\usepackage{wrapfig}
\usepackage{algorithm}
\usepackage{algpseudocode}
\usepackage{booktabs}

\usepackage{bookmark}                   %

\hypersetup{
  colorlinks,
  linkcolor={red!50!black},
  citecolor={blue!60!green},
  urlcolor={blue!60!green}
}
\usepackage{enumitem}

\setlist[itemize]{leftmargin=30pt,rightmargin=25pt}

\title{Unlocking Point Processes through Point Set Diffusion}

\iclrfinalcopy

\author{%
  David L\"udke\thanks{Equal contribution}, Enric Rabasseda Raventós\footnotemark[1], Marcel Kollovieh, Stephan G\"unnemann \\
  Department of Informatics \& Munich Data Science Institute \\
  Technical University of Munich, Germany \\
  \{\texttt{d.luedke,e.rabasseda,m.kollovieh,s.guennemann}\}\texttt{@tum.de}
}

\newtheorem{definition}{Definition}
\newtheorem{lemma}{Lemma}
\newtheorem{theorem}{Theorem}
\newtheorem{corollary}{Corollary}
\newenvironment{proof}{\noindent\textbf{Proof.}}{\hfill $\square$\par}
\begin{document}

\maketitle

\input{sections/abstract}
\input{sections/introduction}

\input{sections/background}

\input{sections/method}

\input{sections/experiments}
\input{sections/related_work}
\input{sections/conclusion}

\bibliography{iclr2025_conference}
\bibliographystyle{iclr2025_conference}

\input{sections/appendix}

\end{document}

%% file: math_commands.tex
\usepackage{amsmath,amsfonts,bm}

\makeatletter
\newcommand*{\rsimdots}{%
  \mathrel{%
    \mathpalette\@rsimdots{}% Adopt to math style size via \mathpalette
  }%
}
\newcommand*{\@rsimdots}[2]{%
  \sbox0{$#1\sim\m@th$}%
  \sbox2{$#1\vcenter{}$}% \ht2 is height of the math axis
  \dimen@=.75\ht2\relax % distance dot to math axis
  \sbox2{$#1\cdot\m@th$}% single vertically centered dot
  \sbox2{% two dots above and below the math axis
    \rlap{\raisebox{\dimen@}{\copy2}}%
    \raisebox{-\dimen@}{\copy2}%
  }%
  \sbox2{$#1\rotatebox[origin=c]{-45}{\copy2}$}%
  \rlap{\hbox to \wd0{\hss\copy2\hss}}%
  \copy0 %
}
\makeatother

\def\diff{\mathop{}\!\mathrm{d}}

\def\eqref#1{equation~\ref{#1}}
\def\1{\bm{1}}

\def\eps{{\epsilon}}

\def\vu{{\bm{u}}}

\def\vx{{\bm{x}}}
\def\vy{{\bm{y}}}

\DeclareMathAlphabet{\mathsfit}{\encodingdefault}{\sfdefault}{m}{sl}
\SetMathAlphabet{\mathsfit}{bold}{\encodingdefault}{\sfdefault}{bx}{n}

%% file: sections/abstract.tex
\begin{abstract}
Point processes model the distribution of random point sets in mathematical spaces, such as spatial and temporal domains, with applications in fields like seismology, neuroscience, and economics.
Existing statistical and machine learning models for point processes are predominantly constrained by their reliance on the characteristic intensity function, introducing an inherent trade-off between efficiency and flexibility.
In this paper, we introduce \textsc{Point Set Diffusion}, a diffusion-based latent variable model that can represent arbitrary point processes on general metric spaces without relying on the intensity function.
By directly learning to stochastically interpolate between noise and data point sets, our approach enables efficient, parallel sampling and flexible generation for complex conditional tasks defined on the metric space.
Experiments on synthetic and real-world datasets demonstrate that \textsc{Point Set Diffusion} achieves state-of-the-art performance in unconditional and conditional generation of spatial and spatiotemporal point processes while providing up to orders of magnitude faster sampling than autoregressive baselines.

\end{abstract}

%% file: sections/introduction.tex
\section{Introduction}
\begin{wrapfigure}[19]{r}{0.55\textwidth}
    \vspace{-1.5cm}
    \centering
    \includegraphics[width=\linewidth, trim={1.1cm 2.8cm 10.cm 1.65cm},clip]{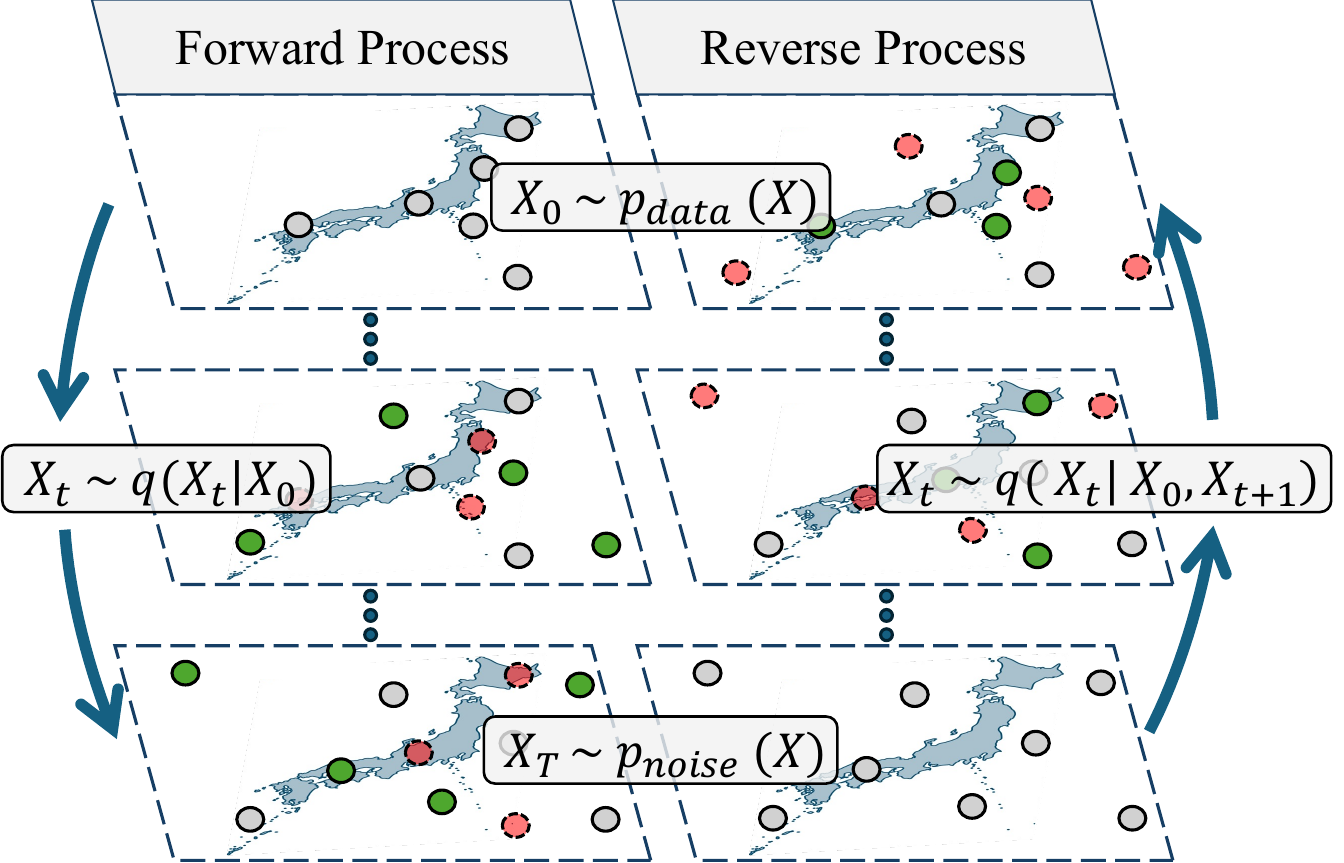}
      \vspace{-0.5cm}
      \caption{Illustration of \textsc{Point Set Diffusion} for earthquakes in Japan. The forward process stochastically interpolates between the original data point set $X_0$ and a noise point set $X_T$, progressively \textcolor{Maroon}{removing} data points and \textcolor{ForestGreen}{adding} noise points. To generate new samples from the data distribution, we approximate the reverse posterior $q(X_t|X_0, X_{t+1})$ and \textcolor{ForestGreen}{add} approximate data points and \textcolor{Maroon}{remove} noise points.}\label{fig:figure1}
\end{wrapfigure}
\looseness=-1
Point processes describe the distribution of point sets in a mathematical space where the location and number of points are random. 
On Euclidean spaces, point processes (e.g., spatial and/or temporal; SPP, STPP, TPP) have been widely used to model events and entities in space and time, such as earthquakes, neural activity, transactions, and social media posts.

Point processes can exhibit complex interactions between points, leading to correlations that are hard to capture effectively \citep{ daley2007introduction}. 
The distribution of points is typically characterized by a non-negative intensity function, representing the expected number of events in a bounded region of space \citep{daley2003introduction}.
A common approach to modeling point processes on general metric spaces is to parameterize an inhomogeneous intensity as a function of space. 
However, this approach assumes independence between points, which limits its ability to capture complex interactions and hinders generalization across different point sets \citep{daley2003introduction, daley2007introduction}.

For ordered spaces like time (STPP, TPP), the predominant approach is to model the conditional intensity autoregressively, where each point is conditioned on the past, allowing for temporal causal dependencies, which can be conveniently captured by state-of-the-art machine learning models \cite{shchur2021neural}. 
While this enables point interactions, these models rely on likelihood-based training and autoregressive sampling, which require integrating the intensity function over the entire space.
Ultimately, this limits possible models, as it either necessitates oversimplified parameterizations that restrict point dependencies and introduce smoothness \citep{ozaki1979maximum,ogata1998space,zhou2023autostpp}, or approximations with amortized inference \citep{zhou2022deepstpp}, numerical \citep{chen2020neural}, or Monte Carlo methods \citep{deep_neyman_scott}. 
Thus, capturing complex point dependencies and sampling from point processes, particularly on general metric spaces, remains an open and challenging problem.

\citet{add_thin} overcame the limitations of the conditional intensity function for temporal point processes by proposing \textsc{Add-Thin}, a diffusion model for TPPs based on the thinning and superposition property of TPPs directly modeling entire event sequences.
In this paper, we generalize this idea to point processes on general metric spaces and derive a diffusion-based latent variable model, \textsc{Point Set Diffusion}, that directly learns to model the stochastic interpolation between a data point set and samples from any noise point process (see \autoref{fig:figure1}).
Furthermore, we show how to generate conditional samples with our unconditional \textsc{Point Set Diffusion} model to solve arbitrary conditioning tasks on general metric spaces.
Our experiments demonstrate that \textsc{Point Set Diffusion} achieves state-of-the-art results on conditional and unconditional tasks for SPPs and STPPs. Our contributions can be summarized as follows:
\begin{itemize}
   \item We derive a diffusion-based latent variable model for point processes on general metric spaces, capturing the distribution of arbitrary point processes by learning stochastic interpolations between data and noise point sets.
    \item Our model supports efficient and parallel sampling of point sets and enables generation for arbitrary conditional tasks defined as binary masks on the metric space.  
    \item We introduce a model-agnostic generative evaluation framework for point process models on Euclidean spaces.
    \item Our method achieves state-of-the-art results for conditional and unconditional generation of SPPs and STPPs while offering orders of magnitude faster sampling.
\end{itemize}

%% file: sections/background.tex
\section{Background}
\looseness=-1

\subsection{Point Processes}
\looseness=-1
A \textit{point process} \citep{daley2003introduction} is a stochastic process where realizations consist of finite sets of randomly located points in a mathematical space. More formally, let $(D, d)$ be a complete, separable metric space equipped with its Borel $\sigma$-algebra $\mathcal{B}$. A point process on $D$ is a mapping \( X \) from a probability space \( (\Omega, \mathcal{A}, \mathcal{P}) \) into \( N^{lf} \), the set of all possible point configurations, such that for any bounded Borel set \( A \subseteq D \), the number of points in \( A \), denoted by \( N(A)\), is a finite random variable.

Given a realization of the point process \( X = \{\vx_i \in D\}_{1 \leq i \leq n} \), where \( n \) is the number of points, the number of points in a region is expressed as the \textit{counting measure} $N(A) = \sum_{i=1}^{n}{\mathbf{1} \{\vx_i \in A\}}$. Here, we assume the point process is simple, i.e., almost surely \( N(\{\vx_i\}) \leq 1 \) for all \( \vx_i \in D \), meaning no two points coincide.
Point processes are commonly characterized by their \textit{intensity function}, which is defined through the following random measure:
\begin{equation}\label{eq:mu_measure}
    A \mapsto \mu(A) \coloneqq \mathbb{E}[N(A)] = \int_A \lambda(\vx) \, \diff \vx,
\end{equation}
where $\mu(A)$ represents the expected number of points in a region \( A \).
Then, a point process is said to have intensity \( \lambda \) if the measure \( \mu \) above has a \textit{density} \( \lambda \) with respect to the Lebesgue measure $\mu(A)$.
Thus, the intensity function \( \lambda(\vx) \) gives the expected number of points per unit volume in a small region of the Borel set \( A \subseteq D \).

As the points in a realization $X$ can exhibit complex correlations, the intensity function can be non-trivial to parameterize. 
For an Euclidean space $\mathbb{R}$ we can specify the Papangelou intensity \citep{daley2003introduction}:
\begin{equation}
    \lambda(\vx) = \lim_{\delta \rightarrow 0} \frac{P\{N(B_\delta(\vx))=1 | C(N(\mathbb{R}\setminus B_\delta(\vx)))\}}{|B_\delta(\vx)|},
\end{equation}
where \(B_\delta(\vx)\) is the ball centered at $x$ with a radius of $\delta$, and \(C(N(\mathbb{R}\setminus B_\delta(\vx)))\) represents the information about the point process outside the ball. If the Euclidean space is ordered, for instance, representing time, the conditioning term would represent the history of all points prior to \( x \).

In general, effectively modeling and sampling from the \textit{conditional intensity} (or related measures, e.g., hazard function or conditional density), for arbitrary metric spaces is generally not possible \citep{daley2003introduction, daley2007introduction}.
This difficulty has led to a variety of simplified parametrizations that restrict the captured point interactions \citep{ozaki1979maximum,zhou2023autostpp,daley2003introduction,daley2007introduction}; discretizations of the space \citep{ogata1998space,osama2019regmethod}; and numerical or Monte Carlo approximations \citep{chen2020neural,deep_neyman_scott}.

In contrast, we propose a method that bypasses the abstract concept of a (conditional) intensity function by directly manipulating point sets through a latent variable model. Our approach leverages the following point process properties:\footnote{We provide a proof of both properties for general Borel sets in \ref{properties}.}

\textit{Superposition:} Given two point processes $N_1$ and $N_2$ with intensities $\lambda_1$ and $\lambda_2$ respectively, we define the superposition of the point processes as $N = N_1 + N_2$ or equivalently $X_1 \bigcup X_2$. Then, the resulting point process $N$ has intensity $\lambda = \lambda_1 + \lambda_2$.
\textit{Independent thinning:} Given a point process $N$ with intensity $\lambda$, randomly removing each point with probability $p$ is equivalent to sampling points from a point process with intensity $(1-p)\lambda$.

\subsection{Diffusion Models}
\looseness=-1
\citet{ho2020denoising} and \citet{sohl2015deep} introduced a new class of generative latent variable models – probabilistic denoising diffusion models.
Conceptually, these models learn to reverse a probabilistic nosing process to generate new data and consist of three main components: a \textit{noising process}, a \textit{denoising process}, and a \textit{sampling procedure}. 
The \textit{noising process} is defined as a forward Markov chain \smash{$q(X_{t+1} | X_t)$}, which progressively noises a data sample \( X_0 \sim p_{\mathrm{data}}(X) \) over $T$ steps, eventually transforming it into a sample from a stationary noise distribution \smash{$X_T\sim p_{\mathrm{noise}}(X)$}. 
Then, the \textit{denoising process} is learned to reverse the noising process by approximating the posterior \smash{$q(X_{t} | X_0, X_{t+1})$ with a model $p_{\theta}(X_{t} | X_{t+1})$}. 
Finally, the \textit{sampling procedure} shows how to generate samples from the learned data distribution \smash{$p_{\theta}(X) = \int p_{\mathrm{noise}}(X_T) \prod_{t=0}^{T-1} p_{\theta}(X_{t} | X_{t+1})\diff X_{1}\dots \diff X_{T}$}.

%% file: sections/method.tex
\section{\textsc{Point Set Diffusion}}
\looseness=-1
In this section, we derive a diffusion-based latent variable model for point sets on general metric spaces by systematically applying the thinning and superposition properties of random sets.
This approach allows direct manipulation of random point sets, avoiding the need for the abstract concept of an intensity function. 
We begin by outlining the forward noising process in \autoref{forward}, which stochastically interpolates between point sets from the generating process and those from a noise distribution.
Subsequently, we demonstrate how to learn to reverse this noising process to generate new random point sets in \autoref{reverse}.
Finally, in \autoref{sampling}, we show how to sample from our unconditional model and generate conditional samples for general conditioning tasks on the metric space.

\subsection{Forward Process}\label{forward}
\looseness=-1
Let $X_0\sim p_{\mathrm{data}}(X)$ be an i.i.d.\ sample from the generating point process, and let $X_T \sim p_{\mathrm{noise}}(X)$ represent a sample from a noise point process. 
We define the forward process as a stochastic interpolation between the point sets $X_0$ and $X_T$ over $T$ steps.
This process is modeled as a Markov chain $q(X_{t+1} | X_t)$, where $X_t$ is the superposition of two random subsets: $X_t^{\mathrm{thin}}\subseteq X_0$ and $X_t^{\eps}\subseteq X_T$.
Specifically, $\forall t: X_t = X_t^{\mathrm{thin}} \bigcup X_t^{\eps}$, where $X_t^{\mathrm{thin}}$ and $X_t^{\eps}$ are independent samples from a \textit{thinning} and a \textit{noise} process, respectively.
We define the \textit{thinning} and \textit{noise} processes given two noise schedules $\{\alpha_t \in (0,1)\}_{t=1}^T$ and $\{\beta_t \in (0,1)\}_{t=1}^T$ as follows:

\begin{figure}
    \centering
    \includegraphics[width=\linewidth, trim={0.5cm 9.0cm 2.0cm 3.5cm}, clip]{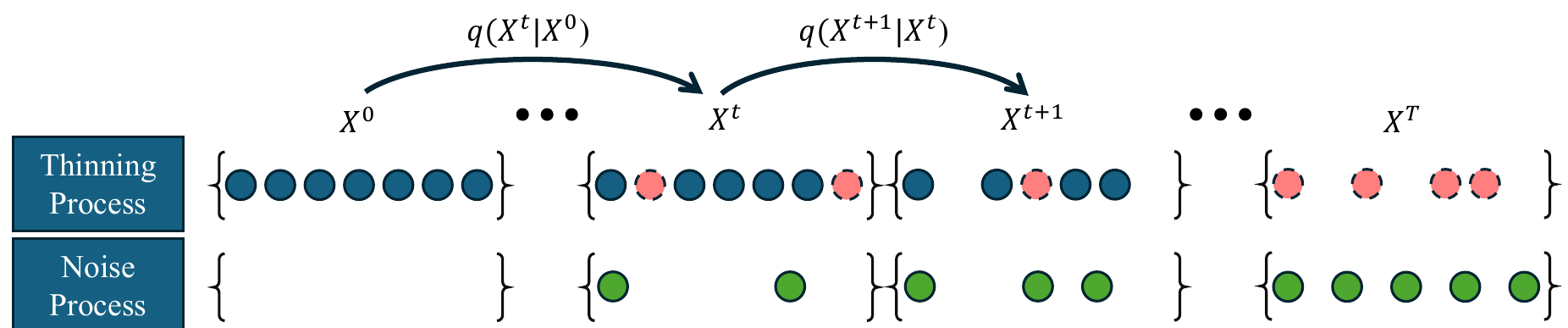}
  \caption{The forward process is a Markov Chain $q(X_{t+1} | X_t)$, that stochastically interpolates a data sample $X_0$ with a noise point set $X_{T}$ over $T$ steps by applying a \textit{thinning} and a \textit{noise} process.}\label{fig:figure2}
\end{figure}
\textbf{Thinning Process:}  This process progressively thins points in $X_0^{\mathrm{thin}}=X_0$, removing signal over time. 
At every step $t+1$, each point $\vx \in X_t^{\mathrm{thin}}$ is independently thinned with probability $1 - \alpha_{t+1}$:
\begin{equation}\label{eq:thin}
    q(\vx \in X_{t+1}^{\mathrm{thin}} | \vx \in X_t^{\mathrm{thin}}) = \alpha_{t+1}.
\end{equation}
Consequently, the thinning defines $n$ independent Bernoulli chains, and the probability of any point $\vx \in X_0$ remaining in $X_t^{\mathrm{thin}}$ is:
\begin{equation}\label{eq:ff_thin}
    q(\vx \in X_t^{\mathrm{thin}} | \vx \in X_0) = \bar{\alpha}_t, 
\end{equation}
where $\bar{\alpha}_t = \prod_{i=1}^n \alpha_i$.
Equivalently, the intensity of the thinned points at step $t$ is given by $\lambda_{t}^{\mathrm{thin}} = \bar{\alpha}_{t} \lambda_{0}$ and the number of remaining points follows a Binomial distribution: $Pr[n_{t} | X_0^{\mathrm{thin}}] = \text{Binomial}(|X_0|, \bar{\alpha}_{t}), \text{where } n_{t}=|X_{t}^{\mathrm{thin}}|$.

\textbf{Noise Process:} This process adds random points $X_T^{\eps} \sim p_{\mathrm{noise}}(X)$ sampled from a noise point process with intensity $\lambda^{\eps}$. 
At step $t+1$, we express $X_{t+1}^{\eps} | X_{t}^{\eps}$ as:
\begin{equation}
     X_{t+1}^{\eps} = X_t^{\eps} \cup X_{t+1}^{\Delta\eps}, \quad \text{where }X_{t+1}^{\Delta\eps} \sim \beta_{t+1} \lambda^{\eps}.
\end{equation}
By the superposition property, the intensity of $X_t^{\eps}$ is $\lambda_t^{\eps} = \bar{\beta}_t \lambda^{\eps},  \text{where } \bar{\beta}_t = \sum_{i=1}^t \beta_i$ and $\bar{\beta}_t\in [0,1]$.
Alternatively, we can view the noise process as a reversed thinning process: we sample $X_T^{\eps} \sim p_{\mathrm{noise}}(X)$ and thin it by $1 - \bar{\beta}_t$ to obtain $X_t^{\eps}$.
Given a noise sample $X_T^{\eps}$, we then find that:
\begin{equation}\label{eq:thin_noise}
    q(\vx \in X_{t}^{\eps} | \vx \in X_T^{\eps}) = \bar{\beta}_{t}.
\end{equation}
Notably, this process is independent of the random point set $X_0$, i.e., $\forall t: q(X_t^{\eps} | X_0) = q(X_t^{\eps}).$

We present a visual depiction of the two forward processes in \autoref{fig:figure2}.
Finally, given that $\forall t: X_t = X_t^{\mathrm{thin}} \bigcup X_t^{\eps}$ it follows that for $\lim_{t\to T} \bar{\alpha}_t=0$ and $\lim_{t\to T}\bar{\beta}_t=1$ the \textit{stationary distribution} is $q(X_T | X_0) = p_{\mathrm{noise}}(X)$, which can be seen by applying the superposition property and finding the intensity of $X_t| X_0$ to be $\bar{\alpha}_t\lambda_{0} + \bar{\beta}_t\lambda^{\eps}$.
To summarize, the forward process gradually removes points from the original point set $X_0 \sim p_{\mathrm{data}}(X)$ while progressively adding points of a noise point set $X_T \sim p_{\mathrm{noise}}(X)$, stochastically interpolating between data and noise.

\subsection{Reverse Process}\label{reverse}
\looseness=-1
To generate samples from our diffusion model, i.e., $X_T \to \cdots \to X_0$, we need to learn how to reverse the forward process by approximating the posterior $q(X_t| X_0, X_{t+1})$ with a model $p_{\theta}(X_t|X_{t+1})$.
We will start by deriving the posterior $q(X_t| X_0, X_{t+1})$ from the forward process $q(X_{t+1}| X_{t})$ and then show how to parameterize and train $p_{\theta}(X_t|X_{t+1})$ to approximate the posterior.

Since the forward process consists of two independent processes (\textit{thinning} and \textit{noise}) and noticing that $X_{t+1}^{\mathrm{thin}} = X_0\bigcap X_{t+1}$ and $X_{t+1}^{\eps} = X_{t+1} \setminus X_0$, the posterior can be derived in two parts:

\textbf{Thinning posterior:} 
Since all points in $X_{t+1}^{\mathrm{thin}}$ have been retained from $t=0$, it follows that $X_{t+1}^{\mathrm{thin}} \subseteq X_{t}^{\mathrm{thin}}$. 
Then for each point in $\vx \in X_0\setminus X_{t+1}^{\mathrm{thin}}$, we derive then posterior using Bayes' theorem, applying \autoref{eq:thin}, \autoref{eq:ff_thin} and the Markov property:
\begin{align}\label{thin_post}
    q( \vx\in X_t^{\mathrm{thin}} | \vx \notin X_{t+1}^{\mathrm{thin}}, \vx \in X_0) &= \frac{q(\vx \notin X_{t+1}^{\mathrm{thin}} | \vx\in X_t^{\mathrm{thin}}) q(\vx\in X_{t}^{\mathrm{thin}} | \vx \in X_0)}{q(x \notin X_{t+1}^{\mathrm{thin}} | \vx \in X_0)}\\
    &= \frac{(1-\alpha_{t+1})  \bar{\alpha}_{t}}{(1-\bar{\alpha}_{t+1})} = \frac{\bar{\alpha}_{t}-\bar{\alpha}_{t+1}}{1-\bar{\alpha}_{t+1}}.
\end{align}
Thus, we can sample $X_{t}^{\mathrm{thin}}$ by superposition of $X_{t+1}^{\mathrm{thin}}$ and thinning $X_0\setminus X_{t+1}^{\mathrm{thin}}$ with \autoref{thin_post}.

\begin{figure}
    \centering
    \includegraphics[width=\linewidth, trim={0.5cm 9.0cm 2.0cm 3.5cm}, clip]{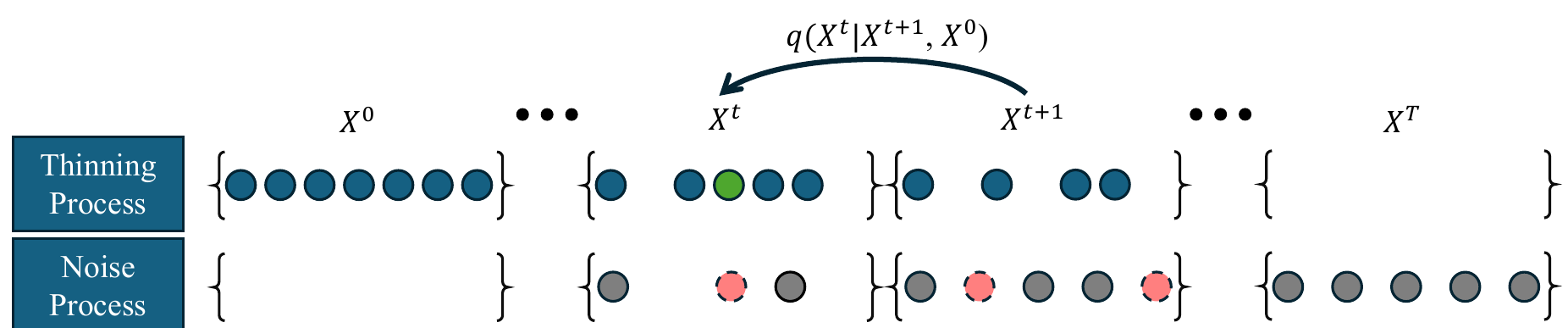}
  \caption{The posterior reverses the stochastic interpolation of $X_0 \to X_T$ of the forward process by adding back thinned points from the thinning process and thinning point added in the noise process.}\label{fig:figure3}
\end{figure}

\textbf{Noise posterior:} Following the reverse thinning interpretation of the noise process, each point in $X_t^{\eps}$ must have been in both $X_{t+1}^{\eps}$ and $X_T^{\eps}$. 
Hence, we derive the posterior for each point in $X_{t+1}^{\eps}$ to still be in $X_{t}^{\eps}$ by following \autoref{eq:thin_noise}, along with the fact that $X_{t}^{\eps}$ is independent from $X_0$:
\begin{align}
    q(\vx\in X_t^{\eps} | \vx \in X_{t+1}^{\eps}, \vx \notin X_0) &= \frac{q(\vx \in X_{t+1}^{\mathrm{thin}} | \vx\in X_t^{\mathrm{thin}}) q(\vx\in X_{t}^{\mathrm{thin}})}{q(\vx \in X_{t+1}^{\mathrm{thin}})}\\
    &= \frac{1 \cdot \bar{\beta}_{t}}{(\bar{\beta}_{t+1})}= \frac{\bar{\beta}_{t}}{\bar{\beta}_{t+1}}.
\end{align}
Thus, we can sample $X_{t}^{\eps}$ by thinning $X_{t+1}^{\eps}$ with probability $(1-\frac{\bar{\beta}_{t}}{\bar{\beta}_{t+1}})$.

\paragraph{Parametrization:}
\looseness=-1
Given $X_0$ and $X_T$, the derived posterior can reverse the noising process to generate $X_0$.
However, to generate a new approximate sample $X_0 \sim p_{\mathrm{data}}(X)$, we need to be able to sample from the posterior $q(X_t| X_0, X_{t+1})$ without knowing $X_0$. 
For this reason we approximate the posterior with a model \smash{$p_{\theta}(X_t| X_{t+1})$}, where we choose \smash{$p_{\theta}(X_t| X_{t+1}) = \int q(X_t| \widetilde{X}_0, X_{t+1}) p_{\theta}(\widetilde{X}_0| X_{t+1}) \diff \widetilde{X}_0$} and training a neural network \smash{$p_{\theta}(\widetilde{X}_0| X_{t})$} to approximate $X_0 | X_{t+1}$ for each $t+1$. 

To effectively train this model, we have to condition our model $p_{\theta}(\widetilde{X}_0| X_{t})$ on $X_t$. 
We propose to embed the points $\vx\in X_{t}$ permutation invariant with a Transformer encoder with full attention and apply a sinusoidal embedding to embed $n=|X_t|$ and $t$. 
Then, to probilistically predict $X_0 | X_t$, we make use of the following case distinction for $X_t^{\mathrm{thin}} = X_0 \bigcap X_t$ and $X_0 \setminus X_t$:

\textit{First}, predicting the retained points in $X_t$, i.e., the intersection of $X_0$ and $X_t$, is a binary classification task for which we train a multi-layer-perceptron (MLP) $g_{\theta}(\vx \in X_t^{\mathrm{thin}} | X_t, t)$ with binary cross entropy loss $\mathcal{L}_{\mathrm{BCE}}$.
\textit{Second}, the thinned points in $X_0$, i.e., $X_0 \setminus X_t$, is a point set $N$, which can be represented by its counting measure, as a mixture of $n$ Dirac measures:
\begin{equation}
    N = \sum_{i=1}^n \delta_{X_i}.
\end{equation}
In \ref{app:approximation_dirac_deltas}, we prove that any finite mixture of Dirac deltas, such as $N$, can be approximated by an $L^{2}$ function in $L^2(D, \mu)$ for any metric space $D$.
In Euclidean spaces, we approximate the Dirac measure with a mixture of multivariate Gaussian distributions with diagonal covariance matrices. 
Note that the multivariate Gaussian density function is a standard approximation of the Dirac delta function and, as the determinant of a diagonal covariance matrix $\bm{\Sigma} \coloneqq \sigma \bm{I}$ approaches zero, the Gaussian increasingly resembles the Dirac delta (See \autoref{Gaussian}). We parameterize the number of points to sample $n_{\theta}$ and the components of the mixture — weights $\bm{w}_{\theta}$, mean $\bm{\mu}_{\theta}$ and diagonal covariance matrix $\bm{\Sigma}_{\theta}$ — with an MLP $f_{\theta}$ and train it with the negative log likelihood $\mathcal{L}_{\mathrm{NLL}}$.

Lastly, to ensure the expected number of points at any time $t$ throughout the diffusion process is constant, we use $\bar{\alpha}_t = 1-\bar{\beta}_t$ and a noise process with a constant intensity such that $\int_A \lambda^{\eps} = E[N(A)]$ for the bounded Borel set $A$ that represents our domain.

\subsection{Sampling Procedure}\label{sampling}
\textbf{Unconditional sampling:}
Starting from a sample $X_T$ of the noise distribution, we apply our \textsc{Point Set Diffusion} model to sample a new $X_0$ over $T$ steps. We start by sampling $X_T \sim \lambda_{\eps}$ and then for all $t \in (T, \dots,1)$ sample $\widetilde{X}_{0} \sim p_{\theta}(X_0 | X_t)$ to subsequently apply the denoising posterior $q(X_{t-1} | \widetilde{X}_{0}, X_t)$ and attain $X_{t-1}$. 
Finally, at step $1$ we sample $\widetilde{X}_{0} \sim p_{\theta}(X_0 | X_1)$. We present the extended sampling algorithm in \autoref{alg:standard_sampling}.

\begin{wrapfigure}[9]{r}{0.4\textwidth}
    \vspace{-0.5cm}
    \centering
    \includegraphics[width=\linewidth, trim={1.7cm 1.8cm 20.3cm 2.1cm},clip]{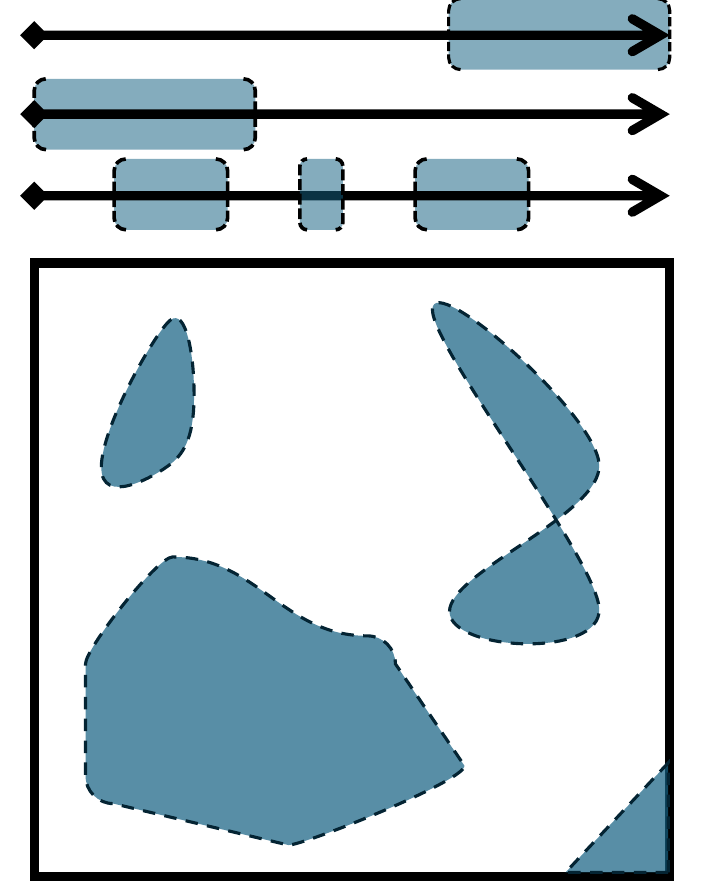}
      \vspace{-0.8cm}
      \caption{Examples of conditioning masks for $\mathbb{R}_{\geq 0}$ and $\mathbb{R}^2$.}\label{fig:figure4}
\end{wrapfigure}
\textbf{Conditional sampling:} Let $C: D \rightarrow \{0,1\}$ be a conditioning mask on our metric space $D$, where we define the masking of a subset $X\subseteq D$ as $C(X) \coloneq \{\vx \in X |  C(\vx)=1\}$ and its complement as $C^{\prime}(X)  \coloneq \{\vx \in X |  C(\vx)=0\}$.
Then, we can leverage our \textsc{Point Set Diffusion} model to conditionally generate random point sets outside the conditioning mask by applying \autoref{algo:conditional_sampling}:
\begin{minipage}{0.57\textwidth}
\vspace{-0.1cm}
    \begin{algorithm}[H]
        \centering
        \caption{Conditional sampling}\label{alg:standard_conditional_sampling}
        \begin{algorithmic}[1]
            \Require $X_0^{c} = C(X_{0})$
            \State $X_T \sim \lambda_{\eps}$
            \For{$t = T, \dots, 1$}
                \State $\widetilde{X}_{0} \sim p_{\theta}(X_0 | X_t)$
                \State  $\widetilde{X}_{t-1} \sim q(X_{t-1} | \widetilde{X}_{0}, X_t)$
                \State $X_{t-1}^{c} \sim q(X_{t-1} | X_0^{c})$
                \State $X_{t-1} = C^{\prime}(\widetilde{X}_{t-1}) \cup C(X_{t-1}^{c})$
            \EndFor
            \State \textbf{return} $C^{\prime}(X_{0})$
    \end{algorithmic}\label{algo:conditional_sampling}
    \end{algorithm}
    \vspace{0.1cm}
\end{minipage}
Thus, following this sampling procedure, we can generate conditional samples for any conditioning mask $C$, where we represent some illustrative conditioning masks for bounded sets on $\mathbb{R}_{\geq 0}$ and $\mathbb{R}^2$ depicting temporal forecasting, history prediction and general imputation tasks in \autoref{fig:figure4}.

%% file: sections/experiments.tex
\section{Experiments}
\looseness=-1
Although point processes are fundamentally generative models, the standard evaluation method relies on reporting the negative log-likelihood (NLL) on a hold-out test set, effectively reducing the evaluation to future single-event predictions for STPPs and TPPS. 
However, this approach presents two key issues. 
\textit{First}, computing the NLL depends on the implementation and parameterization of the (conditional) intensity function and is intractable for many models, necessitating approximations using Monte Carlo methods, numerical integration, or the evidence lower bound (ELBO), complicating a fair comparisons between models.
\textit{Second}, evaluating the likelihood of each point conditioned on ground-truth points does not necessarily reflect how well a model captures the actual data distribution or its ability to perform on complex conditional generation tasks \citep{shchur2021neural}.
To overcome these limitations, we evaluate the generative capabilities of our proposed \textsc{Point Set Diffusion} model by benchmarking it on a range of unconditional and conditional generation tasks for both SPP and STPP. 
The training of our model and the hyperparameters are presented in \ref{app:model_setup}, while all baselines are trained reproducing their reported NLL following their proposed hyperparameters and code.

\subsection{Data}
We follow \citet{chen2021neuralstpp} and evaluate our model on four benchmark datasets: three real-world datasets — \textit{Japan Earthquakes} \citep{data_earthquakes}, \textit{New Jersey COVID-19} Cases \citep{data_covid}, and \textit{Citibike Pickups} \citep{data_citibike} —and one synthetic dataset, \textit{Pinwheel}, based on a multivariate Hawkes process \citep{data_mhp}. 
The pre-processing and splits of the datasets are identical to \citet{chen2021neuralstpp}.

\subsection{Metrics}
To evaluate both unconditional and conditional tasks, we compute distances between point process distributions and individual point sets, assuming the space is normed, and all points are bounded, i.e., $\forall i, \bm{x}_{i} \in [-1,1]^d$. We use the following metrics in our evaluation:

\textbf{Sequence Length (SL):}
To compare the length distribution of point sets, we report the Wasserstein distance between the two categorical distributions. 
For conditional tasks, we compare the length of the generated point set to the ground truth by reporting the Mean Absolute Error (MAE).

\textbf{Counting Distance (CD):}
\citet{xiao2017wasserstein} introduced a Wasserstein distance for ordered TPPs based on Birkhoff's theorem.
We generalize this counting distance to higher-dimensional ordered Euclidean spaces (e.g., STPPs) using the $L_1$ distance:
\begin{equation}\label{eq:counting_distance_spp}
    CD(X, Y) = \frac{1}{d}\sum_{i=1}^{k}{\vert \vert \vx_{i} - \vy_{i} \vert \vert_1} + \sum_{j=k+1}^{l}{\vert \vert U - \vy_{j} \vert \vert_1},
\end{equation}
\begin{wrapfigure}[20]{r}{0.21\textwidth}
    \centering
\vspace{-.84cm}
\includegraphics[width=0.83\linewidth, trim={14cm 1.cm 13.cm 1cm},clip]{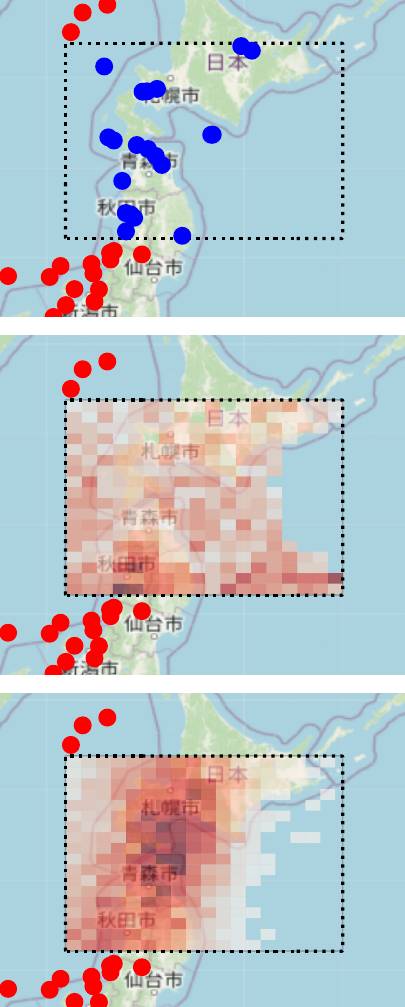}
     \caption{SPP conditioning task: top ground truth, middle \textsc{Regularized Method} and bottom \textsc{Point Set Diffusion}.}\label{fig:figure5}
\end{wrapfigure}
where $X = \{\vx_i\}_{i=1}^{k}$ and $Y = \{\vy_i\}_{i=1}^{l}$ are two ordered samples from a point process on a metric space of dimensionality $d$, i.e. $D \subseteq \mathbb{R}^{d}$. Further, $U \coloneqq (\vu_{1}, \dots, \vu_{d})$ represents the upper bounds of the metric space $D$ along each dimension and we assume, without loss of generality, $l\geq k$.

\textbf{Wasserstein Distance (WD):}
An instance of a Point Process is itself a stochastic process of points in space. Hence, we can compute a distance between two point sets based on the Wasserstein distance on the metric space $D \subseteq \mathbb{R}^{d}$ between the two sets of points.

\textbf{Maximum Mean Discrepancy (MMD) {\normalfont\citep{gretton2012kernel}}:}
The kernel-based statistic test compares two distributions based on a distance metric; we use the WD for SPPs and CD for STPP.

\begin{table}[t!]
\vspace{-0.4cm}
\caption{Density estimation results on the hold-out test set for SPPs, averaged over three random seeds (\textbf{bold} best and \underline{underline} second best).}
\resizebox{\textwidth}{!}{
\begin{tabular}{lcccccccc}
\toprule
& \multicolumn{2}{c}{Earthquakes} & \multicolumn{2}{c}{Covid NJ}
& \multicolumn{2}{c}{Citybike} & \multicolumn{2}{c}{Pinwheel} \\
\cmidrule(lr){2-3}\cmidrule(lr){4-5} \cmidrule(lr){6-7} \cmidrule(lr){8-9}
                    & \multicolumn{1}{c}{SL($\downarrow)$}           & MMD($\downarrow$)           & \multicolumn{1}{c}{SL($\downarrow$)}             & MMD($\downarrow$)            & \multicolumn{1}{c}{SL($\downarrow$)}           & MMD($\downarrow$)           & \multicolumn{1}{c}{SL($\downarrow$)}           & MMD($\downarrow$)           \\
                    \hline
\textsc{Log-Gaussian Cox} & $\underline{0.047}$ & $\underline{0.214}$ & $\underline{0.209}$ & $\underline{0.340}$ & $0.104$ & $\underline{0.336}$ & $\textbf{0.017}$ & $\underline{0.285}$  \\
\textsc{Regularized Method} & $2.361$ & $0.391$ & $0.255$ & $0.411$ &$\underline{0.097}$ & $0.342$ &  $\underline{0.039}$ & $0.411$\\ 
\textsc{Point Set Diffusion} & $\textbf{0.038}$ & $\textbf{0.173}$ & $\textbf{0.199}$ & $\textbf{0.268}$ & $\textbf{0.056}$ & $\textbf{0.092}$ & $\textbf{0.017}$ & $\textbf{0.099}$ \\

\bottomrule
\end{tabular}}
\label{unconditional_spp}
\centering
\bigskip
\caption{Conditional generation results on the hold-out test set for SPP, averaged over three random seeds (\textbf{bold} best).}
\resizebox{\textwidth}{!}{\begin{tabular}{lcccccccc}
\toprule
& \multicolumn{2}{c}{Earthquakes} & \multicolumn{2}{c}{Covid NJ}
& \multicolumn{2}{c}{Citybike} & \multicolumn{2}{c}{Pinwheel} \\
\cmidrule(lr){2-3}\cmidrule(lr){4-5} \cmidrule(lr){6-7} \cmidrule(lr){8-9}
                    & \multicolumn{1}{c}{MAE($\downarrow$)}           & WD($\downarrow$)           & \multicolumn{1}{c}{MAE($\downarrow$)}             & WD($\downarrow$)           & \multicolumn{1}{c}{MAE($\downarrow$)}           & WD($\downarrow$)          & \multicolumn{1}{c}{MAE($\downarrow$)}           & WD($\downarrow$)           \\
                    \hline

\textsc{Regularized Method} 
& 30.419  & 0.162  & 16.075  & 0.148  & 7.740  & 0.115  & 3.547  & 0.150\\
\textsc{Point Set Diffusion}  
&$\textbf{4.651}$ &$\textbf{0.106}$ &$\textbf{5.056}$ &$\textbf{0.119}$ &$\textbf{3.498}$ &$\textbf{0.085}$ &$\textbf{2.256}$ &$\textbf{0.122}$\\ 
\bottomrule
\end{tabular}}
\label{conditional_spp}

\end{table}

\subsection{Spatial Point Processes}

We evaluate our model's ability to capture the distribution of spatial point processes (SPP) by benchmarking it against two methods. The first is the widely used \textsc{Log-Gaussian Cox Process} \citep{moller1998lgcp}, a doubly stochastic model that parameterizes the intensity function using a Gaussian process. 
The second is the \textsc{Regularized Method} \citep{osama2019regmethod}, a spatial model that leverages a regularized criterion to infer predictive intensity intervals, offering out-of-sample prediction guarantees and enabling conditional generation.

\textbf{Unconditional Generation (Density Estimation):}
In this experiment, we generate 1,000 unconditional samples from each model and compare their distribution to a hold-out test set using the WD-SL and WD-MMD metrics. 
As shown in \autoref{unconditional_spp}, our \textsc{Point Set Diffusion} model consistently generates samples most closely matching the data distribution across all datasets.
While the baseline models perform reasonably well in capturing the count distributions for most datasets, their reliance on spatial discretization and smoothness properties of the intensity function limit their ability to capture the complex spatial patterns in the data, as reflected by higher WD-MMD scores.

\textbf{Conditional Generation:} To assess \textsc{Point Set Diffusion}'s ability to solve spatial conditioning tasks, we sample 50 random bounding boxes (with widths uniformly sampled between 1/8 and 3/8 of the metric space) for imputation on the hold-out test set, and report the results in \autoref{conditional_spp}. 
The \textsc{Regularized Method} fits a spatial Poisson model with out-of-sample accuracy guarantees and has been shown by \citet{osama2019regmethod} to outperform the \textsc{Log-Gaussian Cox Process} on interpolation and extrapolation tasks. 
However, we find that the \textsc{Regularized Method}'s reliance on predicting a smooth and discretized intensity function conditioned on neighboring areas leads to inaccurate imputations when the adjacent regions contain significantly different numbers of points (see hexagonal discretization structure and smoothness in \autoref{fig:figure5}). 
This issue is exacerbated by not capturing a shared intensity function across point sets, making it difficult for the \textsc{Regularized Method} to handle non-smooth spatial patterns, such as varying inhomogeneous intensities shared across multiple point sets. 
This highlights a core limitation of SPP models that rely on instance-specific intensity functions.

\subsection{Spatio-temporal Point Processes}
For STPPs, we evaluate our model’s ability to capture the point process distribution by benchmarking it against three state-of-the-art STPP models, learning an autoregressive intensity function.
\begin{wrapfigure}[15]{r}{0.5\textwidth}
    \centering
\vspace{-0.45cm}
\includegraphics[width=\linewidth,clip]{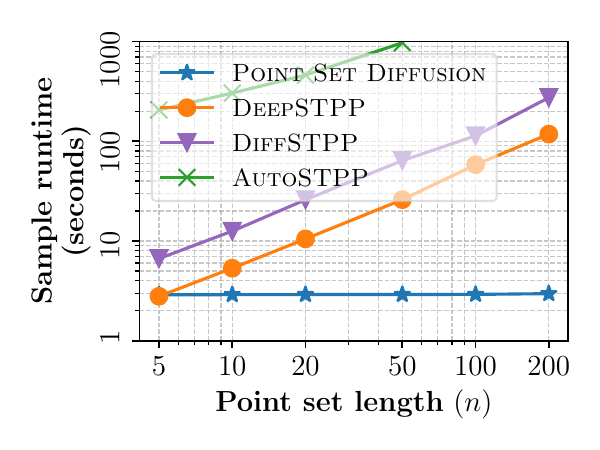}
\vspace{-0.7cm}
     \caption{STPP runtime for sampling $n$ points.}\label{fig:sampling}
\end{wrapfigure}
\textsc{DeepSTPP} \citep{zhou2022deepstpp} uses a latent variable framework to non-parametrically model the conditional intensity based on kernels.
\textsc{DiffSTPP} \citep{yuan2023diffstpp} is based on a diffusion model approximating the conditional intensity.
Lastly, \textsc{AutoSTPP} \citep{zhou2023autostpp} uses the automatic integration for neural point processes, presented by \cite{lindell2021autoint}, to parameterize a generalized spatiotemporal Hawkes model.

\textbf{Sampling Runtime:} We report the median sampling runtime over ten runs generating ten point set of length $n$ on an NVIDIA A100-PCIE-40GB for all STPP models in \autoref{fig:sampling}. \textsc{Point Set Diffusion} achieves a near constant sampling runtime for all point set lengths as it generates all points in parallel, while all autoregressive baselines, given their sequential sampling, show at least a linear relationship between runtime and $n$.

\textbf{Unconditional Generation (Density Estimation):}
We evaluate the performance of each model by comparing the WD-SL and CS-MMD between the hold-out test set and 1,000 samples generated by the trained models, as shown in \autoref{unconditional_stpp}.
Again, the \textsc{Point Set Diffusion} model best captures the distribution of the point process distribution for all datasets.
The autoregressive intensity functions of the baseline models fail to generate point sets that align closely with the data distribution for most datasets, as reflected in the stark differences in the WD-SL and CD-MMD metrics compared to \textsc{Point Set Diffusion}.
While these baselines are trained to predict the next event given a history window, they struggle to unconditionally sample realistic point sets when starting from an empty sequence.
Consequently, this highlights our argument that the standard evaluation based on NLL is insufficient to assess the true generative capacity of point process models.

\textbf{Conditional Generation (Forecasting):}
Forecasting future events based on historical data is a challenging and a fundamental task for STPP models.
To evaluate this capability, we uniformly sampled 50 random starting times from the interval \smash{\([\frac{5}{8}U_{\text{time}}, \frac{7}{8}U_{\text{time}}]\)}, where \(U_{\text{time}}\) is the maximum time, for each point set in the hold-out test set. 
The results are detailed in \autoref{forecasting_stpp}.\footnote{AutoSTPP is not included in this analysis due to its prohibitively slow sampling speed (see \autoref{fig:sampling} and the limitations discussed in \citet{zhou2023autostpp}), which made it impractical to sample the 50 forecast windows for all instances in the test set within a reasonable timeframe.}
The autoregressive baselines, trained to predict the next event based on history, achieve good forecasting results for most datasets, one even surpassing \textsc{Point Set Diffusion} on the Covid NJ dataset. 
Still, our unconditional model outperforms the autoregressive baselines across all other datasets.

\begin{table}[t]
\vspace{-0.4cm}
\centering
\caption{Density estimation results on the hold-out test set for STPP, averaged over three random seeds (\textbf{bold} best and \underline{underline} second best).}
\resizebox{\textwidth}{!}{\begin{tabular}{lcccccccc}
\toprule
& \multicolumn{2}{c}{Earthquakes} & \multicolumn{2}{c}{Covid NJ}
& \multicolumn{2}{c}{Citybike} & \multicolumn{2}{c}{Pinwheel}  \\
\cmidrule(lr){2-3}\cmidrule(lr){4-5} \cmidrule(lr){6-7} \cmidrule(lr){8-9}
                    & \multicolumn{1}{c}{SL($\downarrow$)}           & MMD($\downarrow$)           & \multicolumn{1}{c}{SL($\downarrow$)}             & MMD($\downarrow$)            & \multicolumn{1}{c}{SL($\downarrow$)}           & MMD($\downarrow$)           & \multicolumn{1}{c}{SL($\downarrow$)}           & MMD($\downarrow$)           \\
                    \hline
\textsc{DeepSTPP} & 1.489 & 0.232 & \underline{0.300} & 0.439 & 2.403 & \underline{0.327} & 0.470 & 0.167 \\
\textsc{DiffSTPP} &\underline{0.088} & \underline{0.064} & 0.332 & \underline{0.146} &  0.560 & 0.611 &  0.196 & \underline{0.055} \\
\textsc{AutoSTPP} & 0.169 & 0.472 & 0.390 & 0.382 & \underline{0.556} & 0.412 & \underline{0.111} & 0.105 \\
\textsc{Point Set Diffusion} &\textbf{ 0.042} & \textbf{0.023} & \textbf{0.189} &\textbf{0.043}& \textbf{0.032} & \textbf{0.020} & \textbf{0.023} & \textbf{0.020}  \\

\bottomrule
\end{tabular}}
\label{unconditional_stpp}

\centering

\bigskip
\caption{Forecasting results on the hold-out test set for STPP, averaged over three random seeds (\textbf{bold} best and \underline{underline} second best).}\label{forecasting_stpp}
\resizebox{\textwidth}{!}{\begin{tabular}{lcccccccc}
\toprule
& \multicolumn{2}{c}{Earthquakes} & \multicolumn{2}{c}{Covid NJ}
& \multicolumn{2}{c}{Citybike} & \multicolumn{2}{c}{Pinwheel}  \\
\cmidrule(lr){2-3}\cmidrule(lr){4-5} \cmidrule(lr){6-7} \cmidrule(lr){8-9}
                    & \multicolumn{1}{c}{MAE($\downarrow$)}           & CD($\downarrow$)           & \multicolumn{1}{c}{MAE($\downarrow$)}             & CD($\downarrow$)            & \multicolumn{1}{c}{MAE($\downarrow$)}           & CD($\downarrow$)          & \multicolumn{1}{c}{MAE($\downarrow$)}           & CD($\downarrow$)           \\
                    \hline
\textsc{DeepSTPP}  & \underline{11.743}  & \underline{12.344}  & \textbf{6.912}  & \textbf{8.777}  & 105.494  & 103.397  &  \underline{12.233}  & \underline{10.936}\\ 
\textsc{DiffSTPP} & 16.027  & 17.466  & 18.822  &14.302  & \underline{7.516 }  &  \underline{8.460} & 14.461  & 13.062\\
\textsc{Point Set Diffusion} & 
\textbf{7.407}  & \textbf{10.458}  & \underline{7.293}  & \underline{10.865}  & \textbf{5.928}  & \textbf{7.225}  & \textbf{6.341}  & \textbf{6.437}  \\

\bottomrule
\end{tabular}}
\vspace{-0.5cm}

\end{table}

\subsection{Other conditioning task}

Since the STPP baselines are auto-regressive models, they are limited to forecasting tasks. 
However, our model can generate conditional samples for any conditioning mask $C$ on our metric space. 
To showcase this feature of our model, we present a few visual examples of complex conditioning tasks in \autoref{fig:figure6}.

%% file: sections/related_work.tex
\section{Related Work}
Since large parts of the real-world can be effectively captured by Euclidean spaces, point processes have mainly been defined on spatial and temporal dimensions, represented by an Euclidean space.
Hence, for this discussion of the related work, we will focus on unordered and ordered point processes on Euclidean spaces, mainly SPPs, TPPs and STPPs.
For completeness, we want to mention traditional parametric point processes defined on manifolds, such as determential point processes \citep{berman2008determinantal,katori2022local} and cluster point processes \citep{bogachev2013cluster}.

\textbf{Unordered Point Processes (SPP):}
Modeling a permutation-invariant intensity for unordered point sets that captures complex interactions while remaining efficient for sampling is challenging \citep{daley2007introduction}, seemingly limiting the development of machine-learning-based models for SPPs.
Classical models like the Poisson Point Process \citep{kingman1992poisson} use either homogeneous or inhomogeneous intensity functions across space. 
More flexible models, such as Cox processes \citep{cox1955some}, and specifically the popular \textit{Log-Gaussian Cox Process} \citep{moller1998lgcp}, extend this by modeling the intensity function through a doubly stochastic process, allowing for flexible spatial inhomogeneity.
A recent approach, the \textit{Regularized Method} by \citet{osama2019regmethod}, parameterizes a spatial Poisson process on a hexagonal grid with splines, offering out-of-sample guarantees.
However, these methods often rely on spatial discretization and simple parametric forms and some require separate intensity estimates for each point set, limiting their ability to capture the underlying distribution across different samples \citep{daley2007introduction,osama2019regmethod}.

 \begin{figure}[t]
    \centering
\includegraphics[width=0.8\linewidth, trim={3.cm 1.0cm 2.8cm 1.5cm},clip]{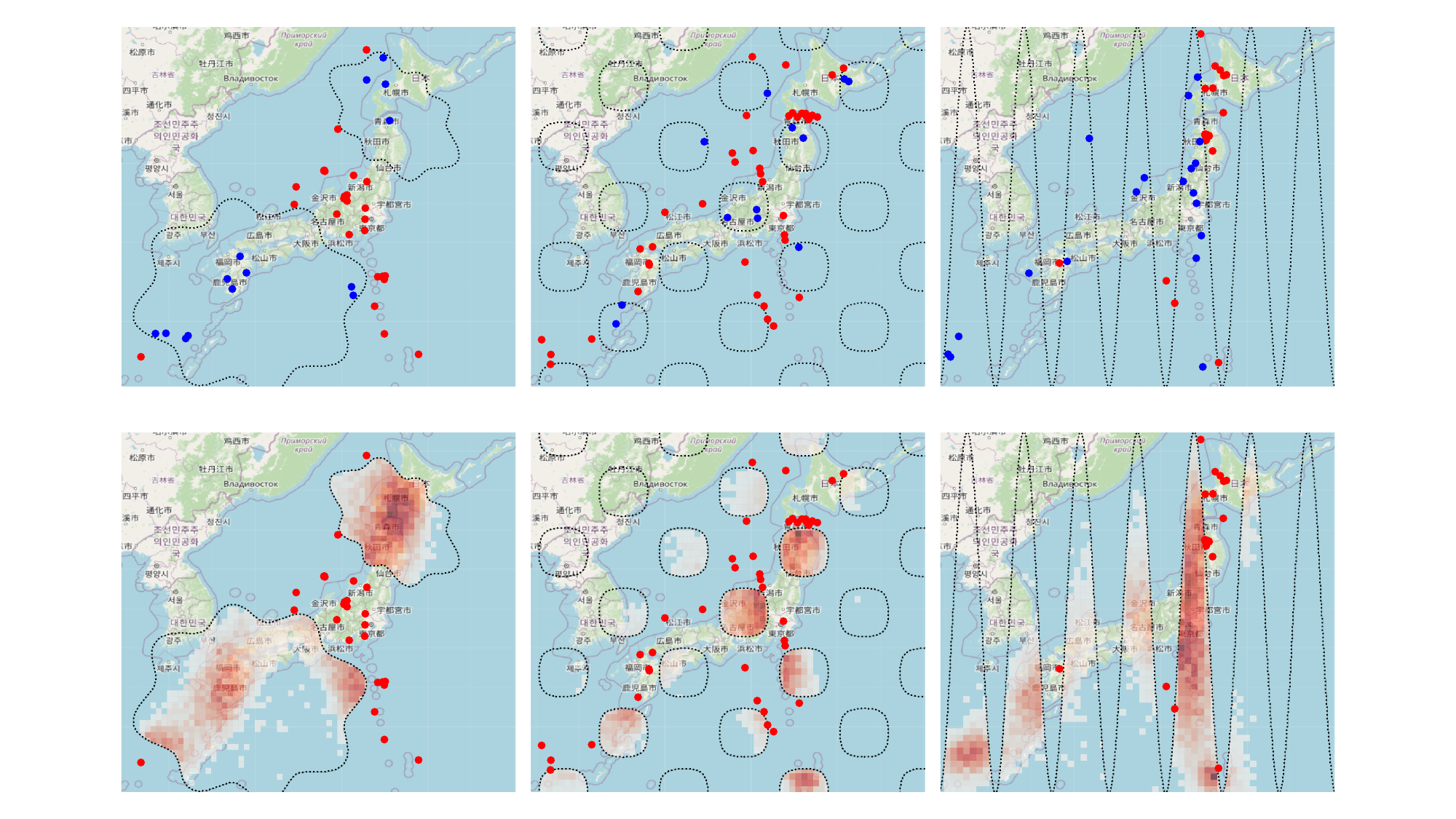}
      \caption{Complex spatial conditioning tasks solved with \textsc{Point Set Diffusion}: Top \textcolor{red}{condition} and \textcolor{blue}{ground truth} data, bottom density plots for predictions.}\label{fig:figure6}
\end{figure}

\textbf{Ordered point processes (TPP and STPP):}
The causal ordering of time enables the parametrization of a conditional intensity, which classically is being modeled with parametric functions, where the Hawkes Process \citep{hawkes1971spectra} is the most widely used model and captures point interaction patterns like self-excitation.
Given the sequential nature of ordered point process a variety of Machine Learning based approaches for TPPs and STPPs have been proposed (see \citet{shchur2021neural} for a review on neural TPPs). 
Where, recurrent neural network- \citep{du2016recurrent,shchur2019intensity} and transformer-based encoders \citep{zhang2020self,zuo2020transformer, chen2020neural} are leveraged to encode the history and neurally parameterized Hawkes \citep{zhou2023autostpp,zhang2020self,zuo2020transformer}, parametric density functions \citep{du2016recurrent,shchur2019intensity}, mixtures of kernels \citep{okawa2019deep,soen2021unipoint,zhang2020cause,zhou2022deepstpp}, neural networks \citep{omi2019fully,zhou2023autostpp}, Gaussian diffusion \citep{linexploring,yuan2023diffstpp} and normalizing flows \citep{chen2020neural,shchur2020fast} have been proposed to (non)-parametrically decode the conditional density or intensity of the next event. 

\textbf{Differences to \textsc{Add-Thin} {\normalfont\citep{add_thin}}:}
Since our method is closely related to \textsc{Add-Thin}, we want to highlight their key methodological differences.
While \textsc{Add-Thin} proposed to leverage the thinning and superposition properties to define a diffusion process for TPPs, \textsc{Point Set Diffusion} generalizes this idea to define a diffusion-based latent variable model for point processes on general metric spaces.
In doing so, we disentangle the superposition and thinning to attain two independent processes to allow for more explicit control and define the diffusion model independent of the intensity function as a stochastic interpolation of point sets.
Furthermore, \textsc{Add-Thin} has to be trained for specific conditioning tasks, while we show how to condition our unconditional \textsc{Point Set Diffusion} model for arbitrary conditioning tasks on the metric space.
Lastly, \textsc{Point Set Diffusion} and its parametrization is agnostic to the ordering of points, making it applicable to model the general class of point processes on any metric space, including for example SPPs.

%% file: sections/conclusion.tex
\section{Conclusion}
To model general point processes on metric spaces, we present \textsc{Point Set Diffusion}, a novel diffusion-based latent variable model.
We derive \textsc{Point Set Diffusion} as a stochastic interpolation between data point sets and noise point sets governed by the thinning and superposition properties of random point sets.
Thereby, we attain a very flexible, unconditional Point Process model that can be conditioned for arbitrary condition masks on the metric space and allows for efficient and parallel sampling of entire point sets without relying on the (conditional) intensity function.
In conditional and unconditional experiments on synthetic and real-world SPP and STPP data, we demonstrate that \textsc{Point Set Diffusion} achieves state-of-the-art performance while allowing for up to orders of magnitude faster sampling.

With \textsc{Point Set Diffusion}, we have presented a novel set modeling approach and would be interested to see how future work explores it's limitations on other (high-dimensional) metric (e.g., Riemannian manifolds), topological and discrete spaces with potential applications extending beyond traditional point sets including but not limited to natural language and graphs.

\newpage

%% file: sections/appendix.tex
\newpage

\appendix

\bookmarksetupnext{level=part}
\pdfbookmark{Appendix}{appendix}

\section{Appendix}

\subsection{Sampling Algorithm}
\begin{minipage}[r]{.5\textwidth}
\begin{flushright}
    \begin{algorithm}[H]
        \centering
        \caption{Sampling}\label{alg:standard_sampling}
        \begin{algorithmic}[1]
            \State $X_T \sim \lambda_{\eps}$
            \For{$t = T, \dots, 1$}
                \State $\widetilde{X}_{t}^{thin} \sim g_{\theta}(\vx \in X_t^{thin} | X_t, t)$
                \State $\widetilde{X}_{0}\setminus X_t \sim f_{\theta}(X | X_t, t)$
                \State $\widetilde{X}_{0} = (\widetilde{X}_{0}\setminus X_t) \cup \widetilde{X}_{t}^{thin}$
                \State  $X_{t-1} \sim q(X_{t-1} \mid \widetilde{X}_{0}, X_t)$
            \EndFor
            \State \textbf{return} $X_{t-1}$
        \end{algorithmic}
    \end{algorithm}\label{alg:sampling}
\end{flushright}
\end{minipage}

\subsection{Point Process Properties}\label{properties}
The thinning and superposition properties have been proved by other works for different versions of point processes. For completeness and generality, we prove them for a general Borel set $A$. To apply these proofs for SPPs consider $A \subseteq \mathcal{S}$, where $\mathcal{S}$ is a metric space in $\mathbb{R}^{d}$ and for STPPs consider $A \subseteq [0, T] \times \mathcal
S$, where $T > 0$.

\paragraph{Superposition:}
\begin{proof}
    It is straightforward to obtain the superposition expectation measure from Equation \ref{eq:mu_measure}: 
    \begin{equation}
        \mu(A) = \mathbb{E}[N(A)] = \mathbb{E}[N_{1}(A) + N_{2}(A)] = \mathbb{E}[N_{1}(A)] + \mathbb{E}[N_{2}(A)] = \mu_{1}(A) + \mu_2(A).
    \end{equation} 
    \noindent Then, every point process has an intensity of $\lambda_1$ and $\lambda_2$ for each of the expectation measures $\mu_1$ and $\mu_2$, respectively. Therefore, taking the right-hand side of Equation \ref{eq:mu_measure}, we obtain the following intensity function for the superposition of point processes: 
    \begin{equation}
        \mu(A) = \mu_1(A) + \mu_2(A) = \int_{A}{\lambda_1(x)dx} + \int_{A}{\lambda_2(x)dx} = \int_{A}{\lambda_1(x) + \lambda_2(x)dx}.
    \end{equation} 
    \noindent This states that the density function for expectation measure $ \mu$ is $\lambda \coloneqq \lambda_1 + \lambda_2$, and concludes the proof for the superposition property of intensities for point processes. 
\end{proof}

\paragraph{Thinning:}
\begin{proof}
    For this property, we need to assume that the singletons are simple, so we can only have one point at each position: $N(\{x\}) \leq 1$; these point processes are called \textit{simple}. Simple point processes can be represented as a sum of Dirac measures at the random points $X_i \in \mathcal{S}$: 
    \begin{equation}
        N = \sum_{i}{\delta_{X_{i}}}.
    \end{equation}
    The previous assumption on singletons makes the sum above a finite sum. If $Z_i \in \{0,1\}$ are Bernoulli random variables with a success probability $p$ we can define a random thinning process as the superposition of the following point processes: 
    \begin{equation}\label{eq:thinning_pp_1}
        N_1 = \sum_{i}{Z_i \delta_{X_{i}}}.
    \end{equation} \\
    \begin{equation}\label{eq:thinning_pp_2}
        N_2 = \sum_{i}{(1-Z_i) \delta_{X_{i}}}.
    \end{equation}\\
    \noindent Since there are only two options for the Bernoulli random variable, it holds that the superposition of the point processes defined in Equations \ref{eq:thinning_pp_1} and \ref{eq:thinning_pp_2} are equivalent to the original point process, i.e., $N = N_1 + N_2$. \\\\
    Given that all $Z_i \sim Bern(p)$ are i.i.d., we obtain a conditional probability distribution on the thinned point process $N_1(A) | N(A) = n \sim Binom(n,p).$ And by the law of total expectation, we derive: 
    \begin{equation}
        \mu_1(A) = \mathbb{E}[N_1(A)] = \mathbb{E}\big[\mathbb{E}[N_1(A) | N(A)] \big] = \mathbb{E}[N(A)p] = \mu(A)p. 
    \end{equation} 
    \noindent We can write the terms of the equation before in terms of the intensity measure of the point process: 
    \begin{equation}
        \mu_1(A) = p \cdot \mu(A) = p \int_A{\lambda(x) dx} = \int_A{p \lambda(x) dx}.
    \end{equation} 
    \noindent Hence, the equation above implies that the intensity of the new point process $N_1$, which keeps the points of the original point process $N$ with probability $p$, is $p\lambda$. \\\\
    By the property of superposition, since $N = N_1 + N_2$, then $\lambda = p\lambda + (1-p)\lambda$. Therefore, the intensity of the point process $N_2$, 
    containing the thinned points, is $(1-p)\lambda$. \\\\
    This proves that, in the opposite case, when removing points with probability $p$ from a given point process with intensity $\lambda$, the intensity of the point process with the points kept after thinning is $(1-p)\lambda$. 
\end{proof}

\subsection{Approximation of Mixture of Dirac Delta Functions by $L^2$-Functions} \label{app:approximation_dirac_deltas}

\begin{definition}[Dirac delta function]
Let $(D, d, \mu)$ be a general metric space equipped with a measure $\mu$. A Dirac delta function $\delta_{\bm{x}}$ at a point $\bm{x} \in D$ is defined as a distribution such that for any test function $f$:
\begin{equation}\label{eq:defdiracdelta}
    \int_{D}{f(\bm{y}) \delta_{\bm{x}}(\bm{y})d\mu(\bm{y})} = f(\bm{x}).
\end{equation}
\end{definition}
\begin{theorem}\label{th:diracdeltaL2}
    Let $f_M(\bm{y})$ be a finite mixture of Dirac deltas:
    \begin{equation}
        f_{M}(\bm{y}) = \sum_{i=1}^n w_i \delta_{\bm{x}_i}(\bm{y}),
    \end{equation}
    where $\bm{x}_1, \dots, \bm{x}_n \in D$ are points in the metric space, and $w_i \in \mathbb{R}$ are weights associated with each Dirac delta function. Then, this finite mixture of Dirac deltas $f_M$ can be approximated by $L^{2}$ functions in $L^2(D, \mu)$.
\end{theorem}

\begin{proof}
    We use a sequence of smooth functions that approximate each Dirac delta in the mixture and then show that this approximation converges in the $L^2$-norm.
    
    Firstly, we show how to approximate Dirac delta functions. Let us consider a family of smooth functions $\phi_\epsilon(\bm{x})$ (such as bump functions or mollifiers) that approximate the Dirac delta function $\delta_{\bm{x}}$ as $\epsilon \to 0$. These functions $\phi_\epsilon(\bm{x} - \bm{x}_i)$ are supported near $\bm{x}_i$ and satisfy:
    \begin{equation}
        \lim_{\epsilon \to 0} \phi_\epsilon(\bm{x} - \bm{x}_i) = \delta_{\bm{x}_i}(\bm{x}).
    \end{equation}
    In particular, for any test function $f$, we have:
    \begin{equation}
        \int_{D} {f(\bm{y}) \phi_\epsilon(\bm{y} - \bm{x}_i) d\mu(\bm{y})} \to f(\bm{x}_i) \quad \text{as} \quad \epsilon \to 0.
    \end{equation}
    Hence, $\phi_\epsilon(\bm{x} - \bm{x}_i)$ has a similar property as the one of Dirac deltas given in Equation \ref{eq:defdiracdelta} and serves as an approximation of the Dirac delta $\delta_{\bm{x}_i}(\bm{x})$ for a small $\epsilon$.

    Secondly, we approximate the mixture of Dirac deltas $f_M$ by a function in $L^2(D, \mu)$ using the same $\phi_\epsilon(\bm{x})$-based approximation for each Dirac delta, defining:
    \begin{equation}
        f_\epsilon(\bm{y}) = \sum_{i=1}^n w_i \phi_\epsilon(\bm{y} - \bm{x}_i).
    \end{equation}
    Each term $\phi_\epsilon(\bm{y} - \bm{x}_i)$ is a smooth approximation of the corresponding Dirac delta $\delta_{\bm{x}_i}(\bm{y})$, and the sum represents the approximation of the entire mixture of Diracs.

    Thirdly, we show that the sequence $f_{\epsilon}$ converges to $f_M$ in the $L^2$-norm, i.e., that:
    \begin{equation}\label{eq:initial_L2_norm}
        \lim_{\epsilon \to 0} \| f_\epsilon - f_M \|_{L^2(D, \mu)} = 0.
    \end{equation}
    Since $f_M$ is a sum of Dirac deltas, it is not directly in $L^2(D, \mu)$, but its approximation $f_\epsilon$ is because each $\phi_\epsilon$ is a smooth function and smooth functions with compact support are in $L^2(D, \mu)$.
    
    We compute now the squared $L^2$ norm of the difference in Equation \ref{eq:initial_L2_norm}:
    \begin{equation}\label{eq:initial_L2_norm_squarted}
    \|f_\epsilon - f_M \|_{L^2(D, \mu)}^2 = \int_D |f_\epsilon(\bm{y}) - f_M(\bm{y})|^2 \, d\mu(\bm{y}).
    \end{equation}
    Note that the squared difference of $f_\epsilon$ and $f_M$ in the above equation will have quadratic and crossed terms. However, we can neglect the crossed terms: $2 \sum_{i < j} w_i w_j \int_D \left( \phi_\epsilon(\bm{y} - \bm{x}_i) - \delta_{\bm{x}_i}(\bm{y}) \right) \left( \phi_\epsilon(\bm{y} - \bm{x}_j) - \delta_{\bm{x}_j}(\bm{y}) \right) \, d\mu(\bm{y})$, since every smooth function $\phi_\epsilon(\bm{y} - \bm{x}_{i})$ is concentrated near $\bm{x}_i$ and terms involving different indices do not contribute to the limit. 
    
    Hence, we can simplify the norm in Equation \ref{eq:initial_L2_norm_squarted} into the sum of the individual terms:
    \begin{equation}
        \| f_\epsilon - f_M \|_{L^2(D, \mu)}^2 = \sum_{i=1}^n \int_D {w_i^2 |(\phi_\epsilon(\bm{y} - \bm{x}_i) - \delta_{\bm{x}_i}(\bm{y}))|^2 d\mu(\bm{y})}.
    \end{equation}
    For every $i$, the term $\int_D {|\phi_\epsilon(\bm{y} - \bm{x}_i) - \delta_{\bm{x}_i}(\bm{y})|^2 d\mu(\bm{y})}$ becomes small as $\epsilon \to 0$, because by construction $\phi_\epsilon(\bm{y} - \bm{x}_i) \to \delta_{\bm{x}_i}(\bm{y})$ in the sense of distributions.
    Thus, by the properties of $\phi_\epsilon$, we conclude that:
    \begin{equation}
        \lim_{\epsilon \to 0} \| f_\epsilon - f_M \|_{L^2(D, \mu)} = 0.
    \end{equation}
\end{proof}

\begin{lemma}\label{lemma:mvn_l2}
    Let $p(\bm{x};\bm{\mu}, \bm{\Sigma})$ be the probability density function (PDF) of a multivariate Gaussian distribution. Then $p \in L^2(\mathbb{R}^d)$.
\end{lemma}
\begin{proof}
The PDF of a multivariate Gaussian distribution in $\mathbb{R}^d$ with mean vector $\bm{\mu} \in \mathbb{R}^d$ and covariance matrix $\bm{\Sigma}$ (which is positive definite) is given by:
\begin{equation}
p(\bm{x};\bm{\mu}, \bm{\Sigma}) = \frac{1}{(2\pi)^{n/2} |\bm{\Sigma}|^{1/2}} \exp \left( -\frac{1}{2} (\bm{x} - \bm{\mu})^T \bm{\Sigma}^{-1} (\bm{x} - \bm{\mu}) \right),
\end{equation}
where $\bm{x} \in \mathbb{R}^d$, $|\bm{\Sigma}|$ is the determinant of the covariance matrix $\bm{\Sigma}$, and $\bm{\Sigma}^{-1}$ is the inverse of the covariance matrix. We show that $\|p\|_{L^2} = \left( \int_{\mathbb{R}^d} |p(\bm{x})|^2 \, d\bm{x} \right)^{1/2}$ is finite.

We need to compute the following integral:
\begin{equation}
    \int_{\mathbb{R}^d} p(\bm{x})^2 \, d\bm{x} = \frac{1}{(2\pi)^n |\bm{\Sigma}|} \int_{\mathbb{R}^d} \exp \left( - (\bm{x} - \bm{\mu})^T \bm{\Sigma}^{-1} (\bm{x} - \bm{\mu}) \right) d\bm{x}.
\end{equation}

To simplify the calculation, we perform a change of variables: $\bm{y} = \bm{\Sigma}^{-1/2} (\bm{x} - \bm{\mu})$. Under this transformation: $(\bm{x} - \bm{\mu})^T \bm{\Sigma}^{-1} (\bm{x} - \bm{\mu}) = \bm{y}^T \bm{y} = \|\bm{y}\|^2$, and the differential $d\bm{x}$ transforms as: $d\bm{x} = |\bm{\Sigma}^{1/2}| \, d\bm{y} = |\bm{\Sigma}|^{1/2} \, d\bm{y}$. Substituting these into the integral, we get:
\begin{equation}
    \int_{\mathbb{R}^d} \exp\left( - (\bm{x} - \bm{\mu})^T \bm{\Sigma}^{-1} (\bm{x} - \bm{\mu}) \right) d\bm{x} = |\bm{\Sigma}|^{1/2} \int_{\mathbb{R}^d} \exp(- \|\bm{y}\|^2) \, d\bm{y} = \pi^{n/2},
\end{equation}

since the remaining integral is a standard Gaussian integral. Thus, the $L^2$-norm integral becomes:
\begin{equation}
    \int_{\mathbb{R}^d} p(\bm{x})^2 \, d\bm{x} = \frac{1}{(2\pi)^n |\bm{\Sigma}|} |\bm{\Sigma}|^{1/2} \pi^{n/2} = \frac{1}{2^{n}\pi^{n/2}|\bm{\Sigma}|^{1/2}}.
\end{equation}

Since the integral is a finite constant, we conclude that the PDF belongs to $L^2(\mathbb{R}^d)$. 

\end{proof}
\begin{corollary}
    Given an Euclidean space $D \subseteq \mathbb{R}^d$, a finite sum of Dirac deltas can be approximated with a mixture of multivariate Gaussian distributions:
    \begin{equation}
        p_{M}(\bm{x}) = \sum_{i=1}^{n}{w_i \cdot \mathcal{N}(\bm{x};\bm{\bm{\mu}_i}, \bm{\Sigma}_i)}.
    \end{equation}
\end{corollary}
\begin{proof}\label{Gaussian}
    Note that we do not show that a mixture of multivariate Gaussian distributions is the best candidate to approximate a finite sum of Dirac deltas. However, note that a multivariate Gaussian distribution is a standard approximation of a Dirac delta function and can, in the limit of a small covariance matrix, i.e. $|\bm{\Sigma}| << 1$, approximate it. 
    
    The aim of this proof is to show that the mixture of Gaussians $p_M$ can be a candidate to approximate the Dirac deltas. From Theorem \ref{th:diracdeltaL2}, this is equivalent to showing that $p_M$ is a $L^{2}$ function.

    To prove this, we need to integrate:
    \begin{equation}
        \int_{D}{p_M(\bm{x})^2 d\bm{x}} =  \sum_{i=1}^k w_i^2 \int_{D} {\mathcal{N}(\bm{x}; \bm{\mu}_i, \bm{\Sigma}_i)^2 d\bm{x}} + 2 \sum_{i \neq j} w_i w_j \int_{D}{\mathcal{N}(\bm{x}; \bm{\mu}_i, \bm{\Sigma}_i) \mathcal{N}(\bm{x}; \bm{\mu}_j, \bm{\Sigma}_j) d\bm{x}}.
    \end{equation}

    On the one hand, Lemma \ref{lemma:mvn_l2} shows that the integrals of the first sum are finite constants. On the other hand, the integrals on the second sum cannot be computed in a closed form, but it is well known that the product decays exponentially as $\|\bm{x}\| \to \infty$ ensuring a finite integral. Therefore, the squared integral is just a sum of finite constants and hence finite. 
\end{proof}
\subsection{Metrics with standard deviations}\label{standard_deviations}
\bigskip

\begin{table}[h]
\vspace{-0.4cm}
\caption{Density estimation results on the hold-out test set for SPPs averaged over three random seeds.}
\resizebox{\textwidth}{!}{
\begin{tabular}{lcccccccc}
\toprule
& \multicolumn{2}{c}{Earthquakes} & \multicolumn{2}{c}{Covid NJ}
& \multicolumn{2}{c}{Citybike} & \multicolumn{2}{c}{Pinwheel} \\
\cmidrule(lr){2-3}\cmidrule(lr){4-5} \cmidrule(lr){6-7} \cmidrule(lr){8-9}
                    & \multicolumn{1}{c}{SL($\downarrow)$}           & MMD($\downarrow$)           & \multicolumn{1}{c}{SL($\downarrow$)}             & MMD($\downarrow$)            & \multicolumn{1}{c}{SL($\downarrow$)}           & MMD($\downarrow$)           & \multicolumn{1}{c}{SL($\downarrow$)}           & MMD($\downarrow$)           \\
                    \hline
Log-Gaussian Cox & 0.047$\pm$ 0.014 & 0.214$\pm$ 0.004 & 0.209$\pm$ 0.011 & 0.340$\pm$ 0.008 & 0.104$\pm$ 0.017 & 0.336$\pm$ 0.014 & 0.017$\pm$ 0.004 & 0.285$\pm$ 0.004 \\
Regularized Method & 2.361$\pm$ 0.064 & 0.391$\pm$ 0.004 & 0.255$\pm$ 0.011 & 0.411$\pm$ 0.003 & 0.097$\pm$ 0.008 & 0.342$\pm$ 0.008 & 0.039$\pm$ 0.003 & 0.411$\pm$ 0.004\\
\textsc{Point Set Diffusion} & 0.038$\pm$ 0.003 & 0.173$\pm$ 0.004 & 0.199$\pm$ 0.002 & 0.268$\pm$ 0.016 & 0.056$\pm$ 0.020 & 0.092$\pm$ 0.020 & 0.017$\pm$ 0.003 & 0.099$\pm$ 0.006\\
\bottomrule
\end{tabular}}

\centering
\bigskip
\caption{Conditional generation results on the hold-out test set for SPP averaged over three random seeds.}
\resizebox{\textwidth}{!}{\begin{tabular}{lcccccccc}
\toprule
& \multicolumn{2}{c}{Earthquakes} & \multicolumn{2}{c}{Covid NJ}
& \multicolumn{2}{c}{Citybike} & \multicolumn{2}{c}{Pinwheel} \\
\cmidrule(lr){2-3}\cmidrule(lr){4-5} \cmidrule(lr){6-7} \cmidrule(lr){8-9}
                    & \multicolumn{1}{c}{MAE($\downarrow$)}           & WD($\downarrow$)           & \multicolumn{1}{c}{MAE($\downarrow$)}             & WD($\downarrow$)           & \multicolumn{1}{c}{MAE($\downarrow$)}           & WD($\downarrow$)          & \multicolumn{1}{c}{MAE($\downarrow$)}           & WD($\downarrow$)           \\
                    \hline

Regularized Method & 30.419 $\pm 0.278$ & 0.162 $\pm 0.003$ & 16.075 $\pm 0.236$ & 0.148 $\pm 0.001$ & 7.740 $\pm 0.173$ & 0.115 $\pm 0.001$ & 3.547 $\pm 0.104$ & 0.150 $\pm 0.003$\\
\textsc{Point Set Diffusion} & 4.651 $\pm 0.159$ & 0.106 $\pm 0.001$ & 5.056 $\pm 0.115$ & 0.119 $\pm 0.001$ & 3.498 $\pm 0.365$ & 0.085 $\pm 0.014$ & 2.256 $\pm 0.037$ & 0.122 $\pm 0.001$\\
\bottomrule
\end{tabular}}

\end{table}

\begin{table}[h]
\vspace{-0.4cm}
\centering
\caption{Density estimation results on the hold-out test set for STPP averaged over three random seeds.}
\resizebox{\textwidth}{!}{\begin{tabular}{lcccccccc}
\toprule
& \multicolumn{2}{c}{Earthquakes} & \multicolumn{2}{c}{Covid NJ}
& \multicolumn{2}{c}{Citybike} & \multicolumn{2}{c}{Pinwheel} \\
\cmidrule(lr){2-3}\cmidrule(lr){4-5} \cmidrule(lr){6-7} \cmidrule(lr){8-9}
                    & \multicolumn{1}{c}{SL($\downarrow$)}           & MMD($\downarrow$)           & \multicolumn{1}{c}{SL($\downarrow$)}             & MMD($\downarrow$)            & \multicolumn{1}{c}{SL($\downarrow$)}           & MMD($\downarrow$)           & \multicolumn{1}{c}{SL($\downarrow$)}           & MMD($\downarrow$)           \\
                    \hline
DeepSTPP & 1.489$\pm$ 0.988 & 0.232$\pm$ 0.046 & 0.300$\pm$ 0.093 & 0.439$\pm$ 0.257 & 2.403$\pm$ 0.884 & 0.327$\pm$ 0.040 & 0.470$\pm$ 0.326 & 0.167$\pm$ 0.032 \\
DiffSTPP & 0.088$\pm$ 0.009 & 0.064$\pm$ 0.024 & 0.332$\pm$ 0.012 & 0.146$\pm$ 0.026 & 0.560$\pm$ 0.045 & 0.611$\pm$ 0.113 & 0.196$\pm$ 0.098 & 0.055$\pm$ 0.005 \\
AutoSTPP & 0.169$\pm$ 0.047 & 0.472$\pm$ 0.275 & 0.390$\pm$ 0.020 & 0.382$\pm$ 0.107 & 0.556$\pm$ 0.077 & 0.412$\pm$ 0.159 & 0.111$\pm$ 0.019 & 0.105$\pm$ 0.005\\
\textsc{Point Set Diffusion} & 0.042$\pm$ 0.003 & 0.023$\pm$ 0.003 & 0.189$\pm$ 0.006 & 0.043$\pm$ 0.003 & 0.032$\pm$ 0.004 & 0.020$\pm$ 0.001 & 0.023$\pm$ 0.003 & 0.020$\pm$ 0.001\\
\bottomrule
\end{tabular}}

\centering

\bigskip
\caption{Forecasting results on the hold-out test set for STPP averaged over three random seeds.}
\resizebox{\textwidth}{!}{\begin{tabular}{lcccccccc}
\toprule
& \multicolumn{2}{c}{Earthquakes} & \multicolumn{2}{c}{Covid NJ}
& \multicolumn{2}{c}{Citybike} & \multicolumn{2}{c}{Pinwheel} \\
\cmidrule(lr){2-3}\cmidrule(lr){4-5} \cmidrule(lr){6-7} \cmidrule(lr){8-9}
                    & \multicolumn{1}{c}{MAE($\downarrow$)}           & CD($\downarrow$)           & \multicolumn{1}{c}{MAE($\downarrow$)}             & CD($\downarrow$)            & \multicolumn{1}{c}{MAE($\downarrow$)}           & CD($\downarrow$)          & \multicolumn{1}{c}{MAE($\downarrow$)}           & CD($\downarrow$)           \\
                    \hline
DeepSTPP & 11.743 $\pm 2.012$ & 12.344 $\pm 1.363$ & 6.912 $\pm 0.990$ & 8.777 $\pm 0.573$ & 105.494 $\pm 28.423$ & 103.397 $\pm 27.577$ & 12.233 $\pm 5.466$ & 10.936 $\pm 3.834$\\
DiffSTPP  & 16.027 $\pm 6.833$ & 17.466 $\pm 5.748$ & 18.822 $\pm 3.381$ & 14.302 $\pm 0.216$ & 7.516 $\pm 1.973$ & 8.460 $\pm 1.773$ & 14.461 $\pm 4.816$ & 13.062 $\pm 3.901$ \\

\textsc{Point Set Diffusion} & 7.407 $\pm 0.285$ & 10.458 $\pm 0.218$ & 7.293 $\pm 0.082$ & 10.865 $\pm 0.130$ & 5.928 $\pm 2.881$ & 7.225 $\pm 2.802$ & 6.341 $\pm 0.108$ & 6.437 $\pm 0.124$\\
\bottomrule
\end{tabular}}
\vspace{-0.5cm}

\end{table}

\subsection{Model Setup}\label{app:model_setup}

\paragraph{Architecture:} The classifier to predict $X_{0} \cap X_{t}$ is a MLP with $3$ layers and ReLU as activation function. The mixture of multivariate Gaussian distribution that approximates $X_{0} \setminus X_{t}$ contains $16$ components, and the parameters are learned with an MLP of $2$ layers and ReLU as an activation function. 

\paragraph{Training:} All models have been trained on an NVIDIA A100-PCIE-40GB. We use \textit{Adam} as the optimizer and a fixed weight decay of $0.0001$ to avoid overfitting. To avoid exploding gradients, we clip the gradients to have a norm lower than $2$.

\paragraph{Hyperparameters:} We use the same hyperparameters for all datasets and types of point processes. We leverage a hidden dimension and embedding size of $32$. For training, we use a batch size is chosen of $128$ and a learning rate of $0.001$.

\paragraph{Early stopping:} We train the models up to $5000$ epochs with early stopping, sampling $100$ sequences from the model and comparing them to the validation split, with WD-SL metric for SPP and the  CD-MMD metric for STPPs.

\subsection{Connecting the applied loss and the ELBO}

In this section we connect our applied loss function and the ELBO to the unknown data distribution. The ELBO of diffusion models \cite{ho2020denoising} is given by: 

\begin{equation}
    \begin{split}
        \mathcal{L}_{ELBO} &= \mathbb{E}_{q}\Big[ D_{KL}\big( q(X_{T} | X_{0})\, ||\, p(X_{T}) \big) - \log{p_{\theta}(X_0 | X_1)} \\ 
        &+ \sum_{t=2}^T{D_{KL} \Big( q(X_{t-1} | X_{0}, X_{t})\, ||\, p_{\theta}(X_{t-1} | X_{t}) \Big)}\Big].
    \end{split}
\end{equation}

The first term in the ELBO is constant as the distributions defining $q(X_{T} | X_{0})$ and $p_{noise}(X_T)$ have not any learning parameters. The second term is minimized as we directly train our model to optimize this likelihood. The last term, consisting on a sum of different KL divergences between two densities, can be defined as:

\begin{equation}
    D_{KL} (q\, ||\, p_{\theta}) = \mathbb{E}_{q}\big[ \log{(q(X_{t-1} | X_{0}, X_{t})} - \log{(p_{\theta}(X_{t-1}|X_{t}))} \big].
\end{equation}

Hence, we minimize the KL divergence by minimizing the only $\theta$-dependent term, on the right-hand side, maximizing the expectation over the log-likelihood $\log{(p_{\theta}(X_{t-1}|X_{t}))}$. As mentioned in \autoref{reverse}, we can divide the reverse process in two parts: the \textit{thinning posterior} and the \textit{noise posterior}.

\paragraph{Thinning posterior.} The component $X_{t-1}^{thin} \coloneqq X_0 \cap X_{t-1}$ is composed of points kept from $X_{t}^{thin}$ and points thinned from  $X_0 \setminus X_{t}^{thin}$. 

For retained points, both the posteriors $q(X_{t-1} | X_{0}, X_{t})$ and $p_{\theta}(X_{t-1}|X_{t})$ are defined by Bernoulli distributions. The cross-entropy $H(q,p)$ can be written by definition as $H(q,p) = H(q) + D_{KL}(q\, ||\, p)$, where $H(q)$ is the entropy. Therefore, minimizing the KL divergence is equivalent to minimizing the (binary) cross entropy. 

For the other points, we sample from the posterior $q(X_{t-1} | X_{0}, X_{t})$ by thinning $X_0 \setminus X_{t}^{thin}$. For this, we approximate first $X_0$ as $X_0 = X_t \cup (X_0 \setminus X_t)$, where the component $X_0 \setminus X_t$ is learned by maximizing the log-likelihood $\mathbb{E}_{q}[\log{(p_{\theta}(X_{t-1}|X_{t}))}].$ Optimizing this term is equivalent to minimizing the negative log-likelihood (NLL) of our model's mixture of multivariate Gaussian, representing the probability density of $X_0 \setminus X_t$ through a mixture model. 

\paragraph{Noise posterior.} The posterior $X_{t-1}^{\epsilon}$ can be derived by thinning $X_t^{\epsilon} = X_t \setminus X_t^{thin}$. In this case the posteriors $q(X_{t-1} | X_{0}, X_{t})$ and $p_{\theta}(X_{t-1}|X_{t})$ directly depend on $X_{t}^{thin}$, which we approximate with the thinning's posterior explained above by minimizing the cross entropy and, equivalently, minimizing the KL divergence.